\documentclass[journal]{IEEEtran}

\usepackage{amsmath,amsfonts}
\usepackage{algorithmic}
\usepackage{algorithm}
\usepackage{array}
\usepackage[caption=false,font=normalsize,labelfont=sf,textfont=sf]{subfig}

\usepackage{textcomp}
\usepackage{stfloats}
\usepackage{url}
\usepackage{verbatim}
\usepackage{graphicx}
\usepackage{cite}
\usepackage{mathptmx} 
\DeclareFontShape{OT1}{ptm}{m}{scit}{<->ssub * ptm/m/it}{}
\usepackage{amssymb}  
\usepackage{diagbox}
\usepackage{booktabs}
\usepackage{multirow}
\usepackage{color}
\usepackage{enumitem}
\usepackage{marvosym}
\begin{document}

\title{AdaTracker: Learning Adaptive In-Context Policy for Cross-Embodiment Active Visual Tracking}

\author{Kui Wu$^{1}$, Hao Chen$^{2}$, Jinzhu Han$^{3}$, Haijun Liu$^{4}$, Churan Wang$^{5}$, Yizhou Wang$^{6}$, Zhoujun Li$^{1}$, Si Liu$^{1}$ and Fangwei Zhong$^{3}$\Letter 
\thanks{Manuscript received: January, 15, 2026; Revised March, 29, 2026; Accepted April, 14, 2026.}
\thanks{This paper was recommended for publication by Editor Jens Kober upon evaluation of the Associate Editor and Reviewers’ comments.
This work was supported by NSFC-62406010, Beijing Natural Science Foundation L252010, the Fundamental Research Funds for the Central Universities, and the State Key Lab of General Artificial Intelligence at Peking University.} 
\thanks{\Letter indicates corresponding author}
\thanks{$^{1}$Kui Wu, Zhoujun Li and Si Liu are with Beihang University, Beijing, China
        {\tt\footnotesize wukui@buaa.edu.cn}}%
\thanks{$^{2} $Hao Chen is with City University of Macau, Macau, China}%
\thanks{$^{3} $Jinzhu Han and Fangwei Zhong are with the School of Artificial Intelligence, Beijing Normal University, Beijing, China
{\tt\footnotesize fangweizhong@bnu.edu.cn}}%
\thanks{$^{4} $Haijun Liu is with Minzu University of China, Beijing, China}%
\thanks{$^{5} $Churan Wang is with Center for Data Science in Clinical Medicine, Peking University Third Hospital, Beijing, China}%
\thanks{$^{6} $Yizhou Wang is with School of Computer Science, Peking University, Beijing, China}
\thanks{Digital Object Identifier (DOI): see top of this page.}
}

\markboth{IEEE Robotics and Automation Letters. Preprint Version. Accepted April, 2026}
{Wu \MakeLowercase{\textit{et al.}}: AdaTracker} 


\maketitle
\begin{abstract}
Realizing active visual tracking with a single unified model across diverse robots is challenging, as the physical constraints and motion dynamics vary drastically from one platform to another.
Existing approaches typically train separate models for each embodiment, leading to poor scalability and limited generalization. To address this, we propose AdaTracker, an adaptive in-context policy learning framework that robustly tracks targets on diverse robot morphologies. Our key insight is to explicitly model embodiment-specific constraints through an Embodiment Context Encoder, which infers embodiment-specific constraints from history. This contextual representation dynamically modulates a Context-Aware Policy, enabling it to infer optimal control actions for unseen embodiments in a zero-shot manner. To enhance robustness, we introduce two auxiliary objectives to ensure accurate context identification and temporal consistency. Experiments in both simulation and the real world demonstrate that AdaTracker significantly outperforms state-of-the-art methods in cross-embodiment generalization, sample efficiency, and zero-shot adaptation.
\end{abstract}

\section{Introduction}
\label{sec:intro}

Embodied Visual Tracking (EVT) is a cornerstone of autonomous agents, requiring agents to actively follow targets in dynamic 3D environments, which has wide-ranging applications in mobile robotics, assistant robotics, and aerial smart photography. 
\begin{figure}[!htb]
    \centering
    \includegraphics[width=1.0\linewidth]{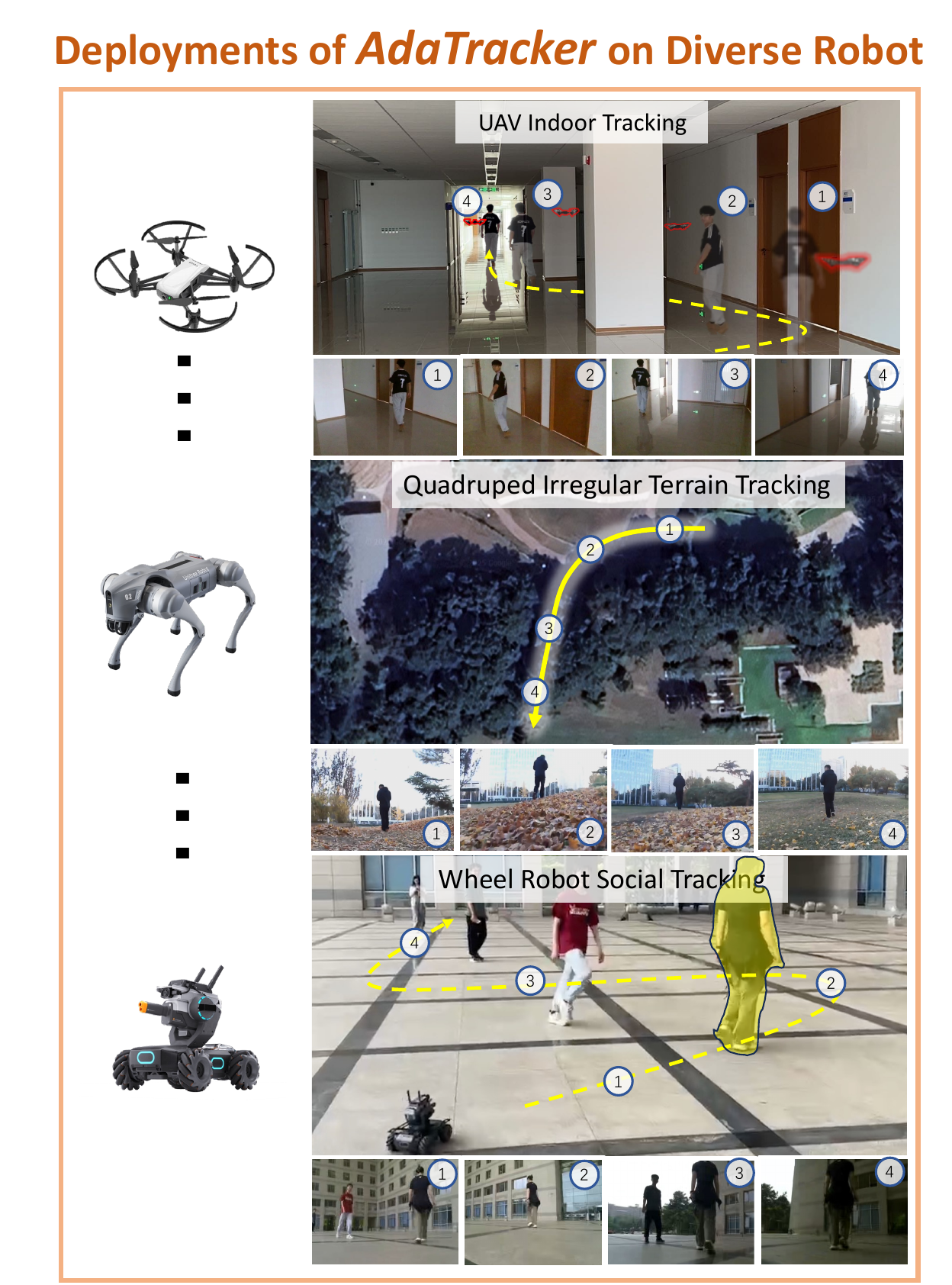}
    \caption{\textbf{AdaTracker} enables a single, unified policy to perform Embodied Visual Tracking (EVT) across \textbf{heterogeneous robotic platforms} with diverse viewpoints and motion dynamics, 
    achieving cross-embodiment generalization and robust 
    real-world tracking without retraining or recalibration.}
    \label{fig:realworld}
    \vspace{-0.5cm}
\end{figure}

Recent EVT methods have progressed from online reinforcement learning ~\cite{zhong2019ad} to offline paradigms and vision-language-action (VLA) frameworks ~\cite{wang2025trackvla}, yet they remain optimized for fixed morphological configurations. Current policies are typically optimized for a 
fixed morphological configuration. Consequently, a policy 
trained for a wheeled robot often fails catastrophically 
when deployed on a high-altitude drone or a trotting 
quadruped. This failure stems from 
coupled perception-control shifts: differing viewpoints and 
speed limits alter action decisions, while differing dynamics 
alter action effects, rendering the observation distribution 
non-stationary across platforms. Such ``embodiment mismatch'' forces robot-specific policy training, multiplying engineering efforts and preventing knowledge reuse across platforms with different viewpoints or dynamics.

To bridge this gap, we present AdaTracker, an adaptive in-context tracking framework designed for zero-shot generalization across heterogeneous robotic platforms. By inferring hidden physical constraints, such as camera height and speed limits from interaction history via the Embodiment 
Context Encoder, AdaTracker conditions a recurrent control policy on the inferred latent context, enabling seamless adaptation without manual fine-tuning or recalibration. Two auxiliary tasks, a supervised context identification objective and a self-supervised temporal consistency objective, further stabilize latent context learning. 
Our main contributions are:

(1) We propose AdaTracker, an adaptive in-context tracking framework, enabling a policy to adaptively control heterogeneous robots.

(2) We extend the standard EVT benchmark to evaluate cross-embodiment generalization. Our method achieves the state-of-the-art (SOTA) performance against strong baselines.

(3)We release an annotated cross-embodiment tracking dataset with 190k steps and a scalable generation pipeline.

(4)We validate the approach through zero-shot transfer experiments on three distinct real-world robots, demonstrating practical cross-embodiment tracking capability.

\section{Related work}
\label{sec:related}
\textbf{Embodied Visual Tracking (EVT)}. The EVT task requires an agent to actively control its motion to keep a target within the field of view. Prior EVT methods relied heavily on Deep Reinforcement Learning (DRL) with online interaction~\cite{luo2019pami, zhang2018coarse, xi2021anti, li2020pose, devo2021enhancing, zhong2019ad, zhong2023rspt}, which is often computationally expensive and time-consuming. To improve efficiency, offline reinforcement learning (Offline RL) methods~\cite{kumar2020conservative, zhong2024empowering} learn policies from pre-collected datasets, avoiding costly online data collection. More recently, Vision-Language-Action (VLA) models such as TrackVLA~\cite{wang2025trackvla} have further advanced EVT by leveraging web-scale data and imitation learning to train robust tracking policies. In this work, we improve cross-embodiment generalization without increasing online interaction cost. We adopt offline reinforcement learning paradigm and condition the control policy on the latent embodiment-context representation during training.

\textbf{Cross-Embodiment Generalization.}
Recent studies explore learning a unified policy~\cite{10611477} that transfers across robots with different dynamics and sensing.
State-based approaches~\cite{sferrazza2024body} often leverage explicit embodiment descriptors and action-space matching, which may be difficult when the target embodiment is unknown, partially observed, or hard to align. Another line of work adapts from limited interaction via in-context RL/trajectory-conditioned policies~\cite{grigsby2024amago,11128272,xu2025less} or context-conditioned dynamics models~\cite{lee2020context}. However, these context representations are typically learned to support prediction or system identification, and are not necessarily suitable as a conditioning signal for embodied visual tracking. In EVT, embodiment changes induce coupled shifts in both visual observations and action effects, making the closed-loop tracking policy highly sensitive to context drift. Consequently, directly encoding context into a latent vector may fail to serve as a reliable \emph{goal representation} for tracking and can destabilize policy execution when used for conditioning. Motivated by this mismatch, we design a \emph{goal-aligned} embodiment context using task-oriented auxiliary objectives tailored to EVT: context identification (camera height prediction and reward prediction) and temporal consistency, then condition the tracking policy on this context~\cite{liu2022goal} to improve closed-loop robustness under distribution shift.

\begin{figure*}
    \centering
    \includegraphics[width=1.0\linewidth]{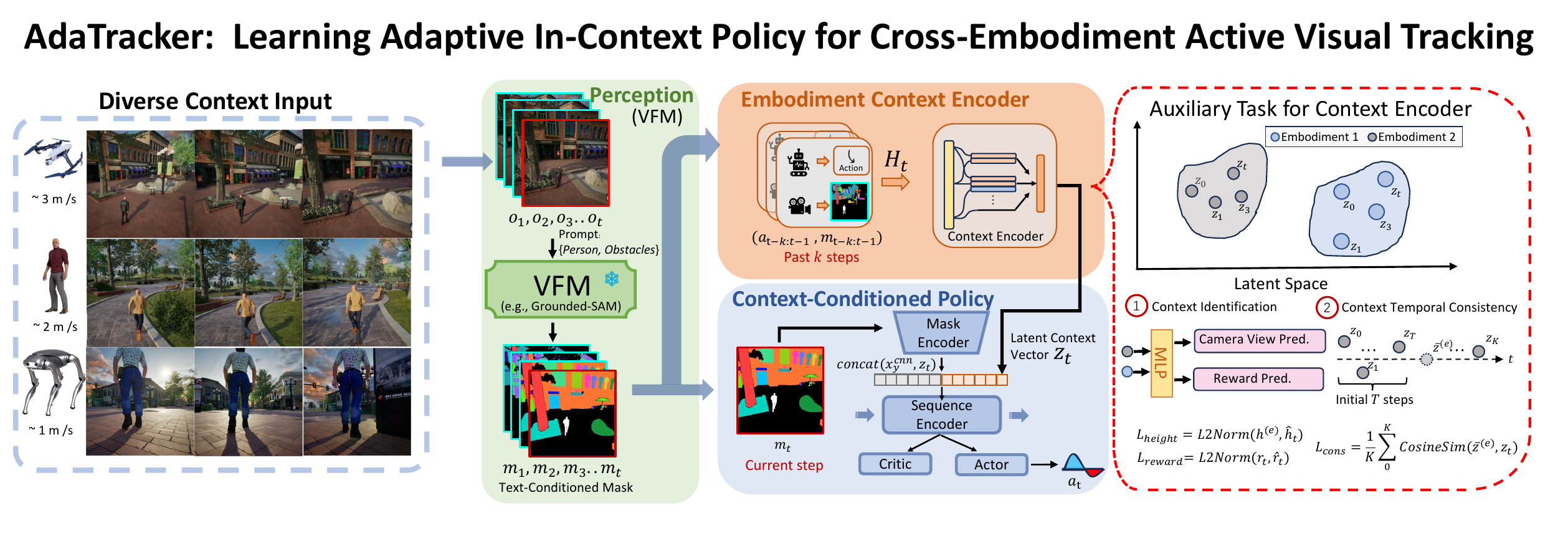}
    \vspace{-1cm}
    \caption{Our framework learns a single, unified policy capable of zero-shot transfer across diverse robot morphologies. The Embodiment Context Encoder (Top Center) infers latent context $z_t$ from historical mask-action pairs via two auxiliary tasks, conditioning the Recurrent Policy (Bottom Center) for adaptive cross-embodiment adaptation.}
    \label{fig:architecture}
    \vspace{-0.5cm}
\end{figure*}

\section{
Adaptive Policy Learning via Self-Aware Context
}
\label{sec:method}
We follow the Embodied Visual Tracking setting~\cite{zhong2021distractor,zhong2024empowering}, where an agent uses ego-centric RGB observations to output actions with a learned policy $\pi_{\theta}$ to keep the target in view. At test time, the policy has no priors of the robot configuration, so it must adapt its behavior to different viewpoints and motion dynamics. We argue that cross-embodiment generalization can be achieved by conditioning the policy on inferred embodiment attributes and action-induced feedback. Accordingly, our framework integrates \emph{self-aware adaptation} via two stages: \textbf{Embodiment Context Encoding} and \textbf{Context-Conditioned Policy Learning}.

\subsection{Embodiment Context Encoding}
Following~\cite{zhong2024empowering}, we use a Visual Foundation Model (VFM;e.g., SAM2~\cite{cheng2023segment}) to produce text-prompted segmentation masks (e.g., ``person, obstacles''), which mitigates visual-induced domain shift. An initialization step automatically selects the target and renders it with a consistent pixel value. We stack masks over the past $K$ steps 
(Fig.~\ref{fig:architecture}) as the context encoding input 
to the RL policy. At deployment, VFM inference is pipelined with the RL controller at $\approx$20\,Hz (Nvidia RTX A3000), meeting the latency requirements of onboard deployment.

Segmentation provides a largely morphology-agnostic semantic input, but cross-embodiment transfer additionally requires awareness of embodiment-dependent dynamics (e.g., camera height, inertia, actuation response). We therefore learn an \emph{Embodiment Context Encoder} that infers a compact latent vector from short histories of masks and actions. At time $t$, we form
$
    H_t = \{(m_\tau, a_\tau)\}_{\tau = t-K}^{t-1}
$,
where $m_\tau$ is the text-conditioned mask and $a_\tau$ is the normalized action. 
$f_\psi$ applies a 
convolutional backbone and a single-layer LSTM~\cite{he2016deep,
lecun1998gradient}, producing $z_t = f_\psi(H_t)$, a 
summary of how the platform responds to actions under its 
viewpoint and dynamic constraints. To encourage $z_t$ to encode embodiment-relevant factors, we train $f_\psi$ with two auxiliary objectives.

\subsubsection{Context identification (supervised)} An auxiliary MLP head $g_\phi$ predicts the immediate reward and camera height: $\hat{r}_t, \hat{h}_t = g_\phi(z_t).$
We use mean-squared error losses 
\begin{equation}
    \mathcal{L}_{\text{reward}} = \mathbb{E}\big[ \big(\hat{r}_t - r_t\big)^2 \big],\mathcal{L}_{\text{height}} = \mathbb{E}\big[ \big(\hat{h}_t - h\big)^2 \big]
\end{equation}

where $r_t$ is the environment reward 
(Section~\ref{sec:task_setting}) and $h$ is the episode-level camera height. Reward prediction ties $z_t$ to action-induced dynamics; height prediction encourages viewpoint sensitivity without explicit calibration.

\subsubsection{Context Temporal Consistency (self-Supervised)} Since embodiment is fixed within an episode, we regularize $z_t$ to stabilize quickly early in the rollout. For episode $e$ of length $T_e$, we define the episode-average context $    \bar{z}^{(e)} = \frac{1}{T_e} \sum_{t=1}^{T_e} z_t^{(e)}$
and penalize deviations of the first $K$ steps from this average:
\begin{equation}
    \mathcal{L}_{\text{cons}} 
    = \mathbb{E}_e \left[1- \frac{1}{K} \sum_{t=1}^{K} CosineSim
    ( \bar z_t^{(e)}, z^{(e)} \right]
\end{equation}
with $K\ll T_e$. This constraint is applied only within each episode, promoting fast identification while allowing different embodiments to occupy distinct regions of the context space. Overall, the representation loss is
$ \mathcal{L}_{\text{rep}} =
     \mathcal{L}_{\text{reward}} +
     \mathcal{L}_{\text{height}} +\mathcal{L}_{\text{cons}}$.
At deployment, we discard $g_\phi$ and use $f_\psi$ to compute $z_t$ online.

\subsection{Context-Conditioned Policy Learning}

\noindent Given the inferred embodiment context $z_t$, we learn a single recurrent policy that adapts its control strategy across heterogeneous platforms by conditioning decision-making on $z_t$.

\subsubsection{Policy architecture.} At time $t$, the current mask $m_t$ is encoded by a convolutional backbone: $x_t^{\text{cnn}} = CNN(m_t).$ In parallel, the context encoder produces $z_t$ from recent mask-action history. We concatenate both features and update a recurrent module: $u_t = \mathrm{concat}\big(x_t^{\text{cnn}}, z_t\big)$ \begin{equation}
    x_t, (h_t, c_t) = \mathrm{LSTM}\big(u_t, (h_{t-1}, c_{t-1})\big).
\end{equation}
$x_t$ serves as the shared input to the actor-critic network, enabling context-conditioned behavior adjustment.

\subsubsection{Optimize policy network via Offline RL.} We adopt Conservative Q-Learning within the Soft Actor-Critic framework (CQL-SAC)~\cite{kumar2020conservative}, comprising critic $Q_\theta$ and actor $V_\phi$.
The CQL-SAC agent comprises a critic network $Q_\theta$, where $Q_\theta$ estimates state–action values, and $V_{\phi}$ represents the control policy, where $\theta$ and ${\phi}$ denote network parameters. In addition, CQL-SAC employs an entropy regularization coefficient $\alpha$ to control the level of exploration. The Q-functions are updated by minimizing the following objective:
\begin{align} 
L_{D}
&=\mathbb{E}_{\mathbf{s}}\left[\log \sum_{\mathbf{a}} \exp (Q^i_\theta(\mathbf{s}, \mathbf{a}))
-\mathbb{E}_{\mathbf{a} \sim \pi_\phi(\mathbf{a} \mid \mathbf{s})}[Q^i_\theta(\mathbf{s}, \mathbf{a})]\right] \nonumber \\
&+ \frac{1}{2} \mathbb{E}_{\mathbf{s}, \mathbf{a}, \mathbf{s}^{\prime}}\left[\left(Q^i_\theta(\mathbf{s}, \mathbf{a}) - \left(r+\gamma E_{a^{\prime} \sim \pi_\phi}\left[Q_{min}\left(s^{\prime}, a^{\prime}\right)\right.\right.\right.\right. \nonumber \\
&\left.\left.\left.\left.-\alpha \log \pi_\phi\left(a^{\prime} \mid s^{\prime}\right)\right]\right)\right)^{2}\right]
\end{align}
where $i \in \{1,2\}$, $Q_{min}=min_{i \in \{1,2\}}Q_\theta^i$ and $Q_\theta^i$ share the same structure. $\gamma$ is the discount factor, $\pi_\phi$ is the policy that derived from the actor network $V_\phi$. 
The first term is the CQL regularization term, which encourages the Q-function to be close to the log-density of the data distribution, and the second term is the standard SAC objective.

The actor is optimized to maximize Q-value while preserving entropy:
\begin{equation}
    \mathcal{L}_{\text{actor}} = \mathbb{E}_{s \sim \mathcal{D}, a \sim \pi_\phi}\left[\alpha \log \pi_\phi(a|s) - Q^{\min}(s,a)\right]
\end{equation}

We jointly optimize the context encoder $f_\psi$ and the policy $\pi_\phi$ on the offline dataset. 
\subsection{Scalable Cross-Embodiment Data Generation} 

To support this architecture and complement the sparsity of cross-embodiment tracking data, we develop an interaction API on top of the previous EVT task simulator ~\cite{qiu2017unrealcv, zhong2025unrealzoo}, enabling scalable cross-embodiment tracking data generation with configurable embodiment parameters. In this paper, we collect 190K step trajectories for training, featuring embodiment-space augmentation across camera heights $\mathcal{H}_{train} = \{0.5m, 1.0m, 1.7m\}$ and velocity limits $\mathcal{V}_{train} \in [0.8, 1.5]$ m/s. 
We select $H_{train}$ and $V_{train}$ to cover representative configurations across common mobile platforms. To evaluate generalization beyond the training distribution, we test on out-of-distribution embodiments (Table~\ref{tab:cross_emb}) with higher viewpoints (up to 3.0 m) and extreme speeds (0.3 m/s and 3.0 m/s).

\section{Experiments}
\noindent Prior EVT works evaluate tracking under fixed robot settings; we instead emphasize \emph{cross-embodiment generalization} and \emph{sim-to-real transfer}, studying: \textbf{(Q1)} zero-shot generalization across viewpoints and kinematic 
limits; \textbf{(Q2)} comparison against classical, online-adaptive, and SOTA baselines; and \textbf{(Q3)} real-world transfer to heterogeneous platforms.

\subsection{Experiment Setup}
\subsubsection{Virtual Environments} We follow the EVT benchmark~\cite{zhong2024empowering} and evaluate on five published simulated scenes: \textit{SimpleRoom, ParkingLot, UrbanCity, UrbanRoad,} and \textit{SnowVillage}, which cover diverse appearance and challenges (e.g., low light, clutter, and uneven terrain).
\subsubsection{Embodiment Configurations} We customize 16 kinds of embodiments by varying camera height and kinematic constraints. Table~\ref{tab:main_results} reports results averaged over all embodiments in each environment, while Table~\ref{tab:cross_emb} breaks down performance across embodiments in \textit{SimpleRoom} to isolate embodiment effects.
\subsubsection{Action space and low-level control}
All embodiments share the same mid-level action interface: the policy outputs velocity commands. We vary the action space across embodiments by imposing embodiment-specific bounds (Table~\ref{tab:cross_emb}). In simulation, an interpolation module converts the commanded velocity into simulator control inputs by smoothly tracking the desired base motion. On real robots, we use the platform's built-in locomotion controller.
Rather than engineering per-embodiment action-interface adapters, we argue that a truly scalable cross-embodiment policy should adapt its own control behavior to varying dynamics via context inference, which is precisely what the Embodiment Context Encoder achieves. The mid-level 
velocity interface thus provides a portable abstraction natively supported across heterogeneous platforms.

\subsubsection{Task Setting} \label{sec:task_setting}
We adopt the standard EVT protocol~\cite{zhong2024empowering,zhong2025unrealzoo}: episodes last up to 500 steps (50 episodes per embodiment); success requires the target within $90^\circ$ FOV and 7.5\,m; failure is declared after 50 consecutive lost steps. The step reward is: $$r_t = 1- \frac{\vert\rho_t - \rho^*\vert}{\rho_{max}} - \frac{\vert\theta_t-\theta^*\vert}{\theta_{max}}$$ where $(\rho_t,\theta_t)$ is the target's relative polar position, $(\rho^*,\theta^*)=(2.5\,\text{m},0^\circ)$ is the desired tracking pose and $(\rho_{\max},\theta_{\max})=(7.5m,\pm 45^\circ)$ denoting the maximum distance and bearing deviations considered in the task.

\subsubsection{Evaluation Metrics} In simulation, we report \textit{Accumulated Reward (AR)} (mean return per episode), \textit{Episode Length (EL)} (mean steps), and \textit{Success Rate (SR)} ~\cite{zhong2024empowering, zhong2025unrealzoo}. In real-world evaluation, ground-truth pose is unavailable, so we report the \textit{Mean Reward (MR)} to replace the AR metric. Specifically, we modify the step reward calculation by using the center and size deviations between the real-time target's bounding box $B_t$ and the initial $B_{\mathrm{init}}$:
$$MR = \frac{1}{L}\sum_{0}^{L} R_{dev}(t), 
R_{dev}=\frac{1}{1+\Delta_c(B_t,B_{\mathrm{init}})+\Delta_s(B_t,B_{\mathrm{init}})}
$$
where
$
\Delta_c(B_t,B_{\mathrm{init}})=\left\|c_t-c_{\mathrm{init}}\right\|_2,\qquad
\Delta_s(B_t,B_{\mathrm{init}})=\left\|[w_t,h_t]-[w_{\mathrm{init}},h_{\mathrm{init}}]\right\|_2,
$
and $c$ denotes the normalized box center while $(w,h)$ denote normalized width and height.

\subsubsection{Baselines.} We compare against methods representing classical control, online adaptation, and recent SOTA approaches: \textbf{(1) PID + Video Tracker}: a conventional pipeline that uses a passive video tracker for target localization (SAM2~\cite{ravi2025sam} + GroundingDINO~\cite{liu2024grounding}) and a PID controller for motion control. \textbf{(2) Online RL}: an online RL agent ~\cite{zhong2019ad} fine-tuned for each embodiment configuration in a sequential (continual) manner, serving as an embodiment-specific adaptation baseline. \textbf{(3) Offline EVT}~\cite{zhong2024empowering}: previous offline-RL SOTA for EVT. \textbf{(4) TrackVLA}~\cite{wang2025trackvla}: recent SOTA vision-language-action model.

\subsection{Results on Simulation}

\begin{table*}[htb]
    \centering
        \centering
        \caption{
        Quantitative results across unseen virtual environments (AR/EL/SR format), averaged over 16 embodiment configurations per environment.
        }

        \resizebox{\textwidth}{!}{
            \begin{tabular}{l|ccccc|c}
            \hline \toprule
            Methods & SimpleRoom & Parking Lot & UrbanCity & UrbanRoad & Snow Village & Mean\\ \midrule
            PID + Video Tracker  & 76/322/0.58  & -87/211/0.34 & -22/289/0.46  &  -168/207/0.28 & -113/218/0.32  &-63/249/0.40 \\ 
            Online RL  &-32/215/0.34& -168/197/0.16& -172/189/0.22 & -178/145/0.14 &-189/169/0.06 & -148/183/0.18\\
            TrackVLA & 128/396/0.62 & -53/319/0.47 & -4/340/0.54 &-78/324/0.47  & 49/303/0.38 &8/336/0.50\\
             Offline EVT & \underline{204}/\underline{451}/\underline{0.76} & \underline{16}/\underline{446}/\underline{0.78} &\underline{34}/\underline{477}/\underline{0.84} &\underline{1}/\underline{446}/\underline{0.79} & \underline{107}/\underline{326}/\underline{0.59}&\underline{72}/\underline{429}/\underline{0.75}  \\
            \textbf{AdaTracker (Ours)}  & \textbf{368/492/0.98} & \textbf{152/488/0.92} & \textbf{212/478/0.94} & \textbf{163/398/0.87} &   \textbf{154/357/0.82}& \textbf{209/443/0.91}\\  \bottomrule
            \end{tabular}
        }
        \label{tab:main_results}
 
\end{table*}

\subsubsection{Comparative Study in Unseen Environments}
We evaluate zero-shot generalization by training on \textit{SimpleRoom} (used for data collection) and testing on four unseen environments spanning indoor to snowy outdoor scenes. Table~\ref{tab:main_results} reports performance averaged over 16 embodiment configurations per environment, measuring cross-embodiment robustness.

Overall, AdaTracker achieves the best results across all environments, demonstrating strong robustness to both environmental shift and embodiment variation. \textbf{PID + Video Tracker} is stable but consistently weaker: although strong perception (e.g., SAM2 ~\cite{ravi2025sam}, GroundingDINO~\cite{liu2024grounding}) improves target localization, a fixed PID controller is hard to tune across embodiments and accumulates errors after brief target losses. \textbf{Online RL} remains sensitive to viewpoint and dynamics changes under cross-embodiment evaluation, and also exhibits worse environmental generalization due to the direct use of RGB observations. Notably, under the sequential continual fine-tuning, we observe performance degradation on previously visited configurations, consistent with catastrophic forgetting and limited transfer across embodiments. \textbf{TrackVLA} is competitive near its training distribution but degrades under out-of-distribution embodiments, consistent with Table~\ref{tab:cross_emb}. \textbf{Offline EVT} attains comparable EL and SR in some environments but substantially lower AR, suggesting weaker preference for high-reward tracking states under distribution shift.

\begin{table*}[htb]
        \centering
        \caption{
        Per-embodiment results in \textit{SimpleRoom} (AR/EL/SR), isolating camera height and target speed across in-distribution and OOD configurations.}

        \label{tab:cross_emb}
        \resizebox{\linewidth}{!}{ 
        \begin{tabular}{lccccc|c}
            \toprule
            \textbf{Method} & \diagbox{Camera Height}{Target Speed}& \textbf{0.5 m/s} & \textbf{1.0 m/s} & \textbf{2.0 m/s} &\textbf{3.0 m/s} &\textbf{   Mean $\pm$ std} \\ 
            \midrule
            \multirow{3}{*}{PID+Video Tracker}
             & 0.3m  &362/496/0.98 & 35/390/0.58&-21/261/0.45		&-44/180/0.35	\\ 
             & 1.0m  &	370/496/0.98 &40/385/0.55&-32/255/0.40 & -33/185/0.38& 76/322/0.58 \\
             & 1.7m   &391/500/1.0	 & 40/395/0.62& -13/270/0.50& -37/190/0.42& $\pm$192/$\pm$213/$\pm$0.41\\
             & 3.0m  &198/302/0.6& 25/375/0.52& -25/275/0.52&-40/195/0.40 \\
            \midrule
            \multirow{3}{*}{Online RL}
             & 0.3m  & -45/221/0.26 & -47/220/0.24 & -35/211/0.32 & -23/260/0.56	  \\ 
             & 1.0m  & -40/205/0.20 & -49/206/0.18 & -44/191/0.36 & -10/260/0.48	  & -32/215/0.34 \\
             & 1.7m   &-37/204/0.22 & -32/197/0.24 & -10/193/0.40 & -15/248/0.52	   & $\pm$33/$\pm$65/$\pm$0.12\\
             & 3.0m  & -49/213/0.24 & -41/208/0.22 & -19/203/0.44 & -16/198/0.54 \\
            \midrule
            
            \multirow{3}{*}{TrackVLA} 
             & 0.3m    & 296/500/1.0 & 379/500/1.0 & 326/500/1.0 & 85/433/0.7 \\ 
             & 1.0m  & 318/500/1.0 & 387/500/1.0 & 268/463/0.76 & 32/383/0.4 & 128/396/0.62 \\
             & 1.7m   & -54/493/0.96 & 89/461/0.86 & 67/352/0.44 & 19/381/0.46 & $\pm$170/$\pm$118/$\pm$0.38\\
             & 3.0m  & -73/198/0.06 & -49/178/0.0 & -9/267/0.22 & -40/227/0.06 \\
            \midrule
            \multirow{3}{*}{Offline EVT} 
             & 0.3m    & 200/491/0.94 & 173/495/0.94 & 66/468/0.82 & 27/488/0.92 \\
             & 1.0m  & 311/496/0.98 & 270/490/0.94 & 111/444/0.6 & 40/351/0.28& 204/451/0.77 \\
             & 1.7m  & 391/500/1.0 &370/500/1.0 & 340/498/0.98 & 277/500/1.0 & $\pm$133/$\pm$76/$\pm$0.30\\
             & 3.0m  &337/484/0.9 & 269/456/0.7 & 91/311/0.28 & 4/259/0.08 \\
             
            \midrule
    
            \multirow{3}{*}{\textbf{AdaTracker (Ours)}} 
             & 0.3m    & 324/500/1.0 & 336/500/1.0 & 240/433/0.96 & 115/494/0.94 \\
            & 1.0m  & 432/500/1.0 & 423/500/1.0 & 385/495/0.96 & 282/498/0.98 & \textbf{368/492/0.98}\\
             & 1.7m   & 464/500/1.0 & 451/500/1.0 & 424/496/0.98 & 369/497/0.96 &$\pm$94/$\pm$17/$\pm$0.03\\
             & 3.0m  & 454/500/1.0 & 433/491/0.96 & 409/488/0.96 & 339/476/0.90 \\
            
            \bottomrule
        \end{tabular}
        }
\end{table*}

\subsubsection{Detailed Analysis of Cross-Embodiment Generalization}
To isolate embodiment effects from environmental factors, we report per-configuration results in \textit{SimpleRoom} (Table~\ref{tab:cross_emb}). The table details the 16 embodiment configurations utilized in Table~\ref{tab:main_results} by sweeping camera height and target speed, including both in-distribution and OOD settings. We test camera heights $\mathcal{H}_{test}=\{0.3,1.0,1.7,3.0\}\,$m and target maximum moving speed up to $3.0\,$m/s, requiring extrapolation to extreme viewpoints and faster dynamics beyond the training distribution.

\textbf{Viewpoint adaptation.}
TrackVLA exhibits a pronounced viewpoint gap: while it performs well at low camera heights, its success rate drops to 0.06 at higher viewpoints. In contrast, AdaTracker maintains consistently strong performance ($SR\geq 0.90$) across all heights, indicating that the context encoder effectively captures perspective-dependent factors and conditions the policy accordingly.

\textbf{Dynamics adaptation.}
Under faster target motion (e.g., $3.0\,\text{m/s}$), TrackVLA exhibits insufficient responsiveness under high-speed target motion and Offline EVT often produces suboptimal control (e.g., $AR\leq 40$), suggesting limited embodiment-aware adaptation. Offline EVT achieves reasonable EL/SR in several configurations, but its markedly lower AR indicates weaker preference for high-reward tracking states, which we attribute to the lack of context inference. AdaTracker remains robust at high speeds (e.g., $SR$ up to 0.98), demonstrating effective adaptation to changing dynamics and kinematic constraints.

Overall, TrackVLA and Offline EVT perform well in configurations close to their training conditions but degrade substantially under embodiment shifts. AdaTracker is consistently strong across the full grid and achieves the best mean AR/EL/SR.

\begin{table}[htb]
    \centering
    \caption{\textbf{Ablation study.} We report mean AR/EL/SR of AdaTracker and variants that remove \emph{Context Identification task}, \emph{Temporal Consistency task}, \emph{entire Embodiment Context Encoding} and the \emph{LSTM structure}.}    
\begin{tabular}[t]{l|c}
                \toprule
               & AR/EL/SR \\ \midrule
               \textbf{Ours}  & \textbf{368/492/0.98} \\ 
               Ours w/o Context Identification & 277/401/0.76\\
               Ours w/o Context Temporal Consistency  & 282/383/0.72\\
               Ours w/o Embodiment Context Encoding& 196/324/0.62  \\
               Ours w/o LSTM  & 172/311/0.60\\ \bottomrule
        \end{tabular}
        \label{tab:ablation}
        \vspace{-0.5cm}
\end{table}

\subsubsection{Ablations}

To evaluate the contribution of each component in AdaTracker, we perform ablations and report the average of three metrics over 16 embodiment configurations (Table~\ref{tab:ablation}). Removing either the \emph{Context Identification} or \emph{Context Temporal Consistency} task degrades performance and often yields poorly tuned control (e.g., overshooting/oscillations). We also remove the entire \emph{Embodiment Context Encoding} and fine-tune the single motion control policy on data pooled across all training embodiments. Without explicit context, the policy becomes unstable under extreme viewpoints and sudden direction change and often fails to re-center the target. 
We attribute this to pooled training inducing ambiguous, multi-modal action requirements across embodiments; without context to disambiguate which high-reward state/action mapping applies, the policy cannot reliably predict appropriate actions.
Finally, ablating the LSTM further reduces long-horizon tracking performance, indicating the importance of temporal aggregation for stable context estimation.

\subsection{Real-world Cross-Embodiment Evaluation} We deploy our policy directly onto three robot platforms without any fine-tuning, including Wheeled Robot, Quadruped robot and Drone. These experiments assess the agent's ability to adapt to real-world dynamics, varying camera perspectives, and environmental disturbances.

\begin{table}[tb]
\caption{Real-world performance comparison of AdaTracker and Offline EVT across three distinct robotic platforms.}
\label{tab:real_world_experiments}
\centering
\resizebox{\columnwidth}{!}{%
\begin{tabular}{l|cccc}
\toprule
\textbf{Platform} & \textbf{Method}  & \textbf{MR} & \textbf{EL} &\textbf{SR}\\ \midrule
Wheeled Robot & Ours &  0.86 $\pm$ 0.04 & 725 $\pm$ 23 & 1.0 \\
             & Offline EVT & 0.82 $\pm$ 0.12 & 726 $\pm$ 22 & 1.0  \\ \hline
Quadruped Robot & Ours &  0.84$\pm$0.06 & 1120 $\pm$ 15 & 1.0 \\
              & Offline EVT & 0.66 $\pm$ 0.26 & 626 $\pm$ 278 & 0.60 \\ \hline
Drone & Ours & 0.83 $\pm$ 0.12 & 1321 $\pm$ 213 & 0.80   \\
      & Offline EVT & 0.42 $\pm$ 0.14 & 454 $\pm$ 28 & 0.00 \\  \bottomrule

\end{tabular}%
}
\vspace{-0.5cm}
\end{table}

\subsubsection{Evaluation Setting}
We evaluate the generalization across three challenge real-world scenarios, including indoor corridor, social crowds, and irregular terrain, as shown in Figure ~\ref{fig:realworld}. During evaluation, the target person follows an S-shaped trajectory, forcing the robot to perform continuous turning and velocity adjustments. We conduct 5 trials for each embodiment.

\subsubsection{Results}
Table~\ref{tab:real_world_experiments} compares AdaTracker with \textbf{Offline EVT}, which is selected as the real-world baseline due to its comparable simulation performance and lightweight inference suitable for on-board deployment. While both methods perform well on the wheeled robot, AdaTracker consistently achieves higher MR and maintains high SR across all platforms. Compared to Offline EVT, AdaTracker explicitly infers embodiment context and conditions the policy on this latent context, enabling adaptive control under platform-dependent viewpoint and dynamics changes. This difference is most evident on the quadruped and UAV. Offline EVT degrades substantially on the quadruped, where irregular terrain induces a larger viewpoint shift when the target moves along a slope; without explicit context inference, the policy tends to produce suboptimal corrective actions and loses stable alignment. 
The gap further widens on the drone, where significant inertial effects and camera shake caused by the aerial viewpoint introduce fast ego-motion and rapid scale/pose changes, further amplifying the mismatch between the training conditions and real-world deployment. As a result, Offline EVT often fails to re-stabilize tracking, leading to substantially lower MR and SR. In contrast, AdaTracker remains robust under these disturbances due to embodiment-context-conditioned control.

\section{Conclusion and Discussion}

Embodiment differences across and within 
robot deployments make per-platform retraining costly. 
AdaTracker addresses this by modeling embodiment variation 
via a compact latent context, enabling a single tracking 
policy to transfer across heterogeneous platforms. Our 
results show that purely RL-based solutions remain 
sensitive to condition changes, while scaling model 
capacity alone is also suboptimal. Future work will explore \emph{conditioned 
RL} to unify diverse mobile tasks into a single 
general-purpose motion control policy via the naturally extensible mid-level velocity interface.



\bibliographystyle{ieeetr}
\bibliography{ref}


\end{document}